\documentclass[11pt,a4paper]{article}
\usepackage[hyperref]{acl2020}
\usepackage{times}
\usepackage{latexsym}

\usepackage{textcomp}

\usepackage{amsmath,amssymb,bm}
\usepackage{graphicx}

\usepackage{microtype}

\aclfinalcopy 
\def\aclpaperid{1887} 

\title{Semi-supervised Formality Style Transfer using Language Model Discriminator and Mutual Information Maximization}

\author{Kunal Chawla \\
  Georgia Institute of Technology \\
  Atlanta, GA \\
  \texttt{kunalchawla@gatech.edu} \\\And
  Diyi Yang \\
  Georgia Institute of Technology \\
  Atlanta, GA \\
  \texttt{dyang888@gatech.edu} \\}

\date{}

\begin{document}
\maketitle
\begin{abstract}
Formality style transfer is the task of converting informal sentences to grammatically-correct formal sentences, which can be used to improve performance of many downstream NLP tasks. In this work, we propose a semi-supervised formality style transfer model that utilizes a language model-based discriminator to maximize the likelihood of the output sentence being formal, which allows us to use maximization of token-level conditional probabilities for training.
We further propose to maximize mutual information between source and target styles as our training objective instead of maximizing the regular likelihood that often leads to repetitive and trivial generated responses. 
Experiments showed that our model outperformed previous state-of-the-art baselines significantly in terms of both automated metrics and human judgement.
We further generalized our model to unsupervised text style transfer task, and achieved significant improvements on two benchmark sentiment style transfer datasets.
\end{abstract}

\section{Introduction}
\begin{figure*}
\includegraphics[width=0.9\linewidth]{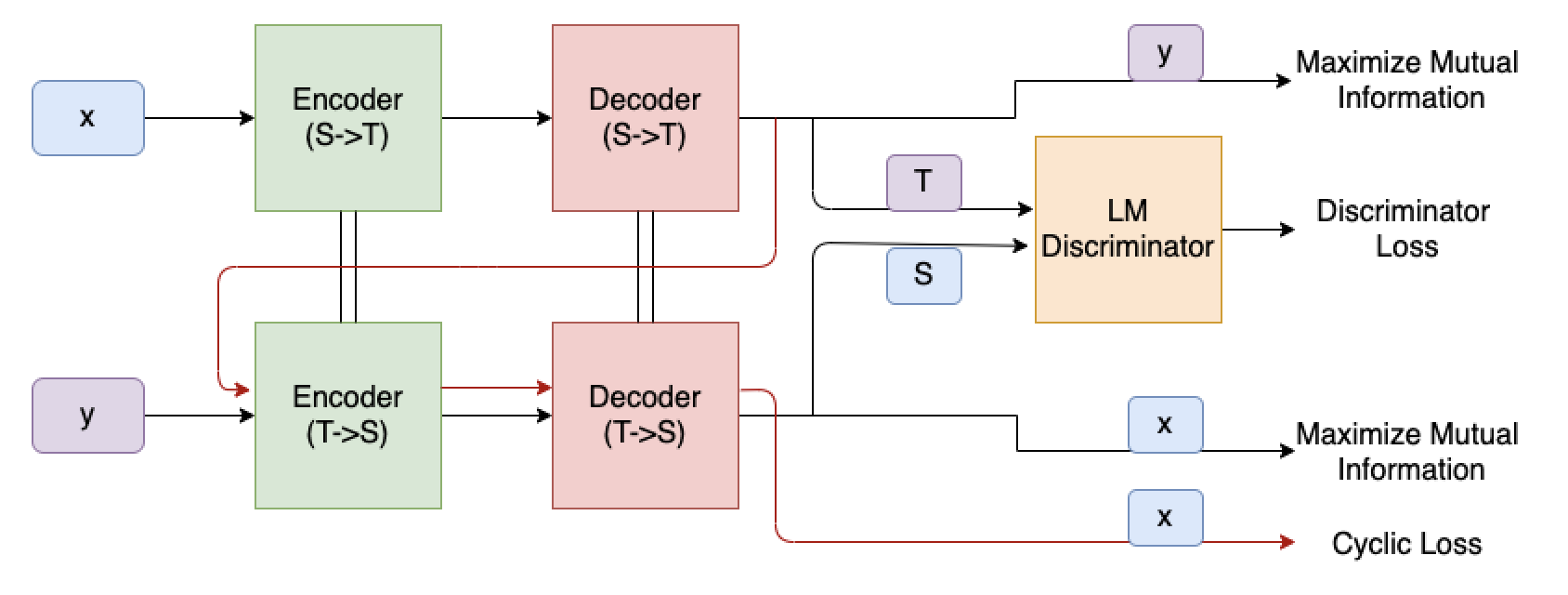}
\caption{Model architecture. Here $(x,y) \in D$ is a (source, target) style sentence pair with same content, and $S$ and $T$ are source and target styles respectively. The parameters for encoder and decoder are shared across forward and backward style transfer directions. The red arrow corresponds to the cyclic reconstruction loss. Cyclic and discriminator losses are trained on $x \in U$, unsupervised class-labeled data.}\label{fig:model}
\end{figure*}

Text style transfer is the task of changing the style of a sentence while preserving the content. It has many useful applications, such as changing emotion of a sentence, removing biases in natural language, and increasing politeness in text~\citep{example1, example2, example3, example4, example5}.

There is a wide availability of ``informal'' data from online sources, yet current Natural Language Processing (NLP) tasks and models could not leverage or achieve good performance for such data due to informal expressions, and grammatical, spelling and semantic errors. Hence, formality style transfer, a specific style transfer task that aims to preserve the content of an informal sentence while making it semantically and grammatically correct, has recently received a growing amount of attention. Some examples are given in Table 1.
\begin{table}[]
    \centering
    \resizebox{0.5 \textwidth}{!} {
    \begin{tabular}{|c |l|}
         \hline
         Informal &  \emph{I flippin' LOVE that movie, sweeeet!} \\
         Formal &  \emph{I truly enjoy that movie.}\\
         \hline
         Informal & \emph{we was hanging out a little.} \\
         Formal &   \emph{We were spending a small amount of time together.} \\
         \hline
    \end{tabular}
    }
    \caption{Examples of (formal, informal) sentence pairs.}
    \label{tab:my_label}
\end{table}

The most widely-used models for formality style transfer are based on a variational auto-encoder architecture, trained on parallel text data of (informal, formal) style sentence pairs with same content \citep{vae1}. However, there is still a lot of inconsistencies between human-generated sentences and outputs of current models, largely due to the limited availability of parallel data. 
In contrast, large amount of data consisting of sentences with just either informal or formal labels is relatively easier to collect. To tackle the training data bottleneck, we propose a semi-supervised approach for formality style transfer, using both human-annotated parallel data and large amount of unlabeled data.

Following the success of Generative Adversarial Nets (GAN) ~\citep{gan}, binary classifiers are often used on the generator outputs in unsupervised text style transfer to ensure that transferred sentences are similar to sentences in the target domain ~\citep{disc1, disc2}. However, ~\citet{lm} showed that using a Language Model instead of a binary classifier can provide stronger, more stable training loss to the model, as it leverages probability of belonging to the target domain for each token in the sentence. We extend this line of work to semi-supervised formality style transfer, and propose to use two language models (one for source style and another for target) to help the model utilize information from both styles for training.

Moreover, style transfer models are usually trained by maximizing $P(y|x)$, where $(x,y)$ is a (informal, formal) sentence pair. Such models tend to generate trivial outputs, often involving high-frequency phrases in the target domain \cite{mmi}.
Building on prior work, to introduce more diversity and connections between the input and output, we propose to maximize mutual information (MMI) between source and target styles, which take into account not only the dependency of output on input, but also the likelihood that the input corresponds to the output. While this has only been done at test-time so far, we extend this approach to train our model with MMI objective.

We evaluate our proposed models that incorporate both the language model discriminators and mutual information maximization on Grammarly Yahoo Answers Corpus (GYAFC) Dataset ~\citep{grammarly}.
Experiments showed that our simple semi-supervised formality style transfer model outperformed state-of-the-art methods significantly, in terms of both automatic metrics (BLEU) and human evaluation.
We further show that our approach can be used for unsupervised style transfer, as demonstrated by significant improvements over baselines on two sentiment style benchmarks: Yelp and Amazon Sentiment Transfer Corpus, where parallel data is not available.  We
have publicly released our code at \url{ https://github.com/GT-SALT/FormalityStyleTransfer}.

\section{Related Works}
\paragraph{Sequence-to-Sequence Models} Text style transfer is often modeled as a sequence-to-sequence (seq2seq) task~\citep{lm,formality,delete}. A classical architecture for seq2seq models is variational autoencoders(VAE) which uses an ``encoder'' to encode the input sentence into a hidden representation, and then uses a ``decoder'' to generate the new sentences~\citep{disc1,disc2, vae1}. Long Short Term Memory(LSTMs)~\citep{lstm}, and more recently, self-attention based CNN architectures ~\citep{transformer} are often used as base architectures for such models. 

Pre-training of the encoders on multiple tasks and datasets has been shown to be effective ~\citep{bert,roberta} in improving performances of individual tasks. These models are often trained with the cross-entropy loss ~\citep{transformer} on the output tokens, or in other words, maximising $P(y|x)$ where $(x,y)$ is a pair of source and target style sentence respectively. ~\citet{mmi} showed that maximising mutual information (MMI) $M(x,y)$ during test-time between the source and target instead can lead to more diverse and appropriate outputs in seq2seq models. Some other works \citep{mmilowerbound2} maximize a variational lower bound on pairwise mutual information. We use a denoising auto-encoder BART \citep{bart} trained with MMI objective.

\paragraph{Semi-Supervised and Unsupervised Style Transfer}
Some approaches like \citet{delete} and \citet{conditionslgan} focus on deleting style-related keywords to make content style-independent. However, other works hypothesize that content and style cannot be separated, and use techniques such as back-translation \citep{backtrans}, cross-projection between styles in latent space \citep{crossprojection}, reinforcement learning-based one step model \citep{reinforcement}, and iterative matching and translation \citep{iterative}. Following \citet{gan}, using a generator along with a style classifier is often used for unsupervised tasks ~\citep{disc1, disc2, unsupbase}. However, recent work suggests ~\citep{lm} that using Language Models instead of CNN discriminators can result in more fluent, meaningful outputs.
Maximizing likelihood of reconstruction of the input from the generated output has been used in both image generation \citep{cyclegan} and text style transfer ~\citep{cycle, reinforcement, cycle2} to improve performance. Motivated by these work, we use language models for our discriminator, and maximize cyclic reconstruction likelihood as part of our training objective.

\paragraph{Formality Style Transfer} Grammarly \citep{grammarly} released a large-scale dataset for Formality Style Transfer, and tested several rule-based and deep neural networks-based baselines. CNN-based discriminators and cyclic reconstruction objective have been used \citep{formality} in a semi-supervised setting. ~\citet{rules} used a combination of original and rule-based processed sentences to train the model. There is also evidence that using multi-task learning \cite{nmtmultitask} and models pretrained on a large scale corpus \cite{rules} improve performance. This work uses a BART model \citep{bart} pretrained on CNN-DM dataset \cite{cnn} for our base architecture.

\section{Method}
This section presents our semi-supervised formality style transfer model. We detail the task and our base architecture in Section 3.1. We add a language model-based discriminator to the model, described in Section 3.2, and explain the maximization of mutual information in Section 3.3. The final architecture for our model is summarized in Section 3.4 and shown in Figure~\ref{fig:model}

\subsection{Formality Style Transfer}
Define $T$ (=``\texttt{formal}'' in our case) as the target style and $S$ (=``\texttt{informal}'') as the source style for the formality style transfer task. Let $D$ be the parallel dataset containing (source, target) style sentence pairs and $U$ be the additional unlabeled data, denoted by $U_S$ for sentences with source style and $U_T$ for sentences with target style.

Our base model is a variational auto-encoder mechanism $G$ that generates sentences of target style. The goal is to maximize $P(y|x;\theta_G)$ where $\theta_G$ are the parameters of the model. This is done by cross-entropy loss over the target sentence tokens and generated output probabilities.
To leverage Maximum Mutual Information objective, as described in Section 3.3, we make the model bi-directional. It can be used to transfer source style to target style as well as target style to source style. Hence, an additional input $c \in \{S,T\}$ is passed to $G$ specifying the style to which the sentence is to be converted. Hence, our objective for base model is to maximize $P(y|x,T; \theta_G)$.

\subsection{Language Model Discriminator}
We add a Language model(LM) based discriminator to the model. It functions as a binary classifier which scores the formality of the output generated by the decoder. It includes two language models trained independently on informal and formal data. The ``score'' of a sentence by a language model is calculated by the product of locally normalized probabilities of each token given the previous tokens. Let $x$ be a sentence from $P$ with label $c$, then
\begin{equation}
\begin{aligned}
LM(x) = \prod_{i=0}^{len(x)} P(x_i | x_{0:i-1}; \theta_{LM})
\end{aligned}
\end{equation}
where $x_i$ are the tokens in $x$ and $\theta_{LM}$ are the parameters of the language model. The softmax-normalized score of the sentence by the language models is interpreted as the classifier score:
\begin{equation}
\begin{aligned}
P(T|x) = \frac{e^{LM_T(x)}}{e^{LM_T(x)} + e^{LM_S(x)}}
\end{aligned}
\end{equation}
The language model discriminator is pre-trained on source and target data from $P$ with the cross entropy loss:
\begin{equation}
\begin{aligned}
\theta_C^* = \text{min}_{\theta_C} \sum_{x \in U} (- \text{log } P(c|x; \theta_C))
\end{aligned}
\end{equation}
where $c$ is the label of $x$, $\theta_C$ are the parameters of the LM discriminator and $\theta_C^*$ are the trained parameters.
The weights are then frozen for the training. A common training objective (\citep{rules, unsupbase}) is to minimize the sum of translation loss $L_{trans}$ and discriminator loss $L_{disc}$, defined as: 
\begin{equation*}
\begin{split}
L_{trans}((x,y) \in D) =& - \text{log }(P(y|x,T; \theta_{G}))) \\
L_{disc}(x \in U_S)  =& - \text{log }(P(T|x,T; \theta_{G}, \theta_C)) \\
L_{disc}(x \in U_T)  =& - \text{log }(P(S|x,S; \theta_{G}, \theta_C)) \\
\end{split}
\end{equation*}

\begin{equation}
\begin{aligned}
\theta_{G}^* =& \text{min}_{\theta_{G}} (\sum_{(x,y) \in D}L_{trans}(x,y) + \sum_{x \in U} L_{disc}(x))
\end{aligned}
\end{equation}

\subsection{Maximum Mutual Information Objective}
As discussed, instead of using usual translation loss which maximizes $ P(y|x; \theta_G)$ and often produces trivial and repetitive content, we chose to maximize pairwise mutual information between the source and the target:
\begin{equation}
\begin{split}
\hat{y} =& \text{argmax}_y \text{log }\frac{P(x,y)}{P(y)P(x)} \\
=& \text{argmax}_y (\text{log } P(y|x) - \text{log } P(y))
\end{split}
\end{equation}
Following \citep{mmi}, we introduce a parameter $\lambda$ ``forward-translation weight'' to generalize the MMI objective and adjust the relative weights of forwards and backwards translation: 
\begin{equation*}
\begin{split}
\hat{y} =& \text{argmax}_y (\text{log } P(y|x) - (1 - \lambda) \text{ log } P(y)) \\
 =& \text{argmax}_y (\lambda \text{ log } P(y|x) + (1- \lambda) \text{ log } P(x|y))
\end{split}
\end{equation*}
The translation loss thus becomes:
\begin{equation}
\begin{split}
L_{trans}((x,y) \in D) = \lambda \; \text{ log }P(y|x,T; \theta_G) \\
 + (1- \lambda) \text{ log }P(x|y,S; \theta_G)
\end{split}
\end{equation}

\subsection{Overall Model Architecture}
Making the model bi-directional also allows us to leverage unsupervised data using cyclical reconstruction loss $L_{cycle}$, which encourages a sentence translated to the opposite style and back to be similar to itself \citep{cycle}. Let $G(x,c)$ be the output of the model for a sentence $x$ with target style $c$. Then
\begin{equation*}
\begin{split}
L_{cycle}(x \in U_S) =& - \text{log }P(x |G(x,T),S; \theta_G)) \\
L_{cycle}(x \in U_T) =& - \text{log }P(x| G(x,S),T; \theta_G))  \\
\end{split}
\end{equation*}
Let $w_{disc}$ and $w_{cycle}$ denote the weights for discriminator and cyclic loss respectively. The overall loss function $L$ for the training step is:
\begin{equation}
\begin{split}
L =& \sum_{(x,y) \in D} L_{trans}(x,y) \\
+ & \sum_{x \in U} (w_{disc} L_{disc}(x) + w_{cycle} L_{cycle}(x))
\end{split}
\end{equation}

\begin{table}
{\begin{tabular}{|l |c c c|}
\hline
 Dataset & Train & Valid & Test \\ 
 \hline
 E\&M & 52595 & 2877 & 1416 \\
 F\&R & 51967 & 2788 & 1432\\
 \hline
 BookCorpus & 214K & - & -\\
 Twitter & 211K & - & -\\
 \hline
 Yelp (Positive) & 270K & 2000 & 500\\
 Yelp (Negative) & 180K & 2000 & 500\\
\hline
 Amazon (Positive) & 277K & 985 & 500\\
 Amazon (Negative) & 278K & 1015 & 500\\
 \hline
\end{tabular}
}
\caption{The statistics of train, validation and test sets of all used datasets.}
\end{table}

\section{Experiments}
\subsection{Dataset}
We used Grammarly's Yahoo Corpus Dataset (GYAFC) ~\citep{grammarly} as our parallel data for supervised training. The dataset is divided into two sub-domains- ``Entertainment and Music'' (E\&M) and ``Family and Relationships'' (F\&R). For the unsupervised data, we crawled Twitter data for informal data, and we used BookCorpus data ~\citep{bookcorpus} for the formal data. In the pre-training step, we train the language model discriminator on the unannotated informal and formal data. The detailed process of the data collection is given in the Appendix. The statistics of  datasets are in Table 2.
\subsection{Pre-processing and Experiment Setup}
The text was pre-processed with Byte Pair Encoding(BPE) ~\citep{bpe} with a vocabulary size of 50,000. For pre-training, we trained the LM Discriminator with the unsupervised data with cross entropy loss. For training, we merged both datasets of GYAFC and used the training objective as described in Section 3.4 to train the model.

We used Fairseq ~\citep{fairseq} library built on top of PyTorch ~\citep{pytorch} to run our experiments. We used BART-large ~\citep{bart} model pretrained on CNN-DM summarization data ~\citep{cnn} for our base encoder and decoder. BART was chosen because of its bidirectional encoder which uses words from both left and right for training, as well as superior performance on text generation tasks. Its training objective of reconstruction from noisy text data fits our task well. We chose the model pre-trained on CNN-DM dataset because of the relevance of the decoder pre-trained on formal words to our task.

Both decoder and the encoder have 12 layers each with 16 attention heads and a hidden embedding size of 1024. We shared the weights for encoder and decoder across the forward and backward translation, using a special input token to the encoder. For the language models, we used a Transformer \citep{transformer} decoder with 4 layers and 8 attention heads per layer.

One NVIDIA RTX 2080 Ti with 11GB memory was used to run the experiments with the max token size of 64. We also used update frequency 4, increasing the effective batch size. Adam Optimizer \citep{adam} was used to train the model, and the parameters learning rate, $\lambda, w_{disc}$ and $w_{cycle}$ were fine-tuned. The model was selected based on perplexity of informal to formal translation on validation data. Beam search (size = 10) was used to generate sentences. A length penalty (= 2.0) was used to reduce redundancy in the output sentence. Further details on model parameters are mentioned in Appendix.

\subsection{Evaluation Metrics}
The result was evaluated with BLEU ~\citep{bleu}. We used word tokenzier and corpus BLEU calculator from Natural Language Toolkit (NLTK) ~\citep{nltk} to calculate the BLEU score. Due to the subjective nature of the task, BLEU does not capture the output of the model well. Hence, we also used human annotations for some of the models. 

Amazon Mechanical Turk was used to evaluate 100 randomly sampled sentences from each dataset of GYAFC. 
To increase annotation quality, we required workers located in US to have a 98\% approval rate and at least 5000 approved HITs for their previous work on MTurk. Each sentence was annotated by 3 workers, who rated each generated sentence using the following metrics, following ~\citep{grammarly}: 
\begin{itemize}
    \item \textbf{Content}: Annotators judge if the source and translated sentence convey the same information on a scale of 1-6: 6: Completely equivalent, 5: Mostly equivalent, 4: Roughly equivalent, 3: Not equivalent but share some details, 2: Not equivalent but on same topic, 1: Completely dissimilar.
    \item \textbf{Fluency}: Workers score the clarity and ease of understanding of the translated sentence on a scale from 1-5: 5: Perfect, 4: Comprehensible, 3: Somewhat Comprehensible, 2: Incomprehensible, 1: Incomplete.
    \item \textbf{Formality}: Workers rate the formality of the translated sentence on a scale of -3 to 3.  -3: Very Informal, -2: Informal, -1: Somewhat Informal, 0: Neutral, 1: Somewhat Formal, 2: Formal and 3: Very Formal. 
\end{itemize}
We also provided detailed definitions and examples to workers, which are described together with annotation interface in Appendix. 
The intra-class correlation was estimated using ICC-2k (Random sample of k raters rate each target) and calculated using Pingouin \citep{pingo} Python package. It varied from 0.521-0.563 for various models, indicating moderate agreement \citep{icc}. We then averaged the three human-provided labels to obtain the rating for each sentence. 

\begin{table*}
\begin{center}
\resizebox{!}{8em}{
\begin{tabular}{|l|c c c c| c c c c|}
\hline
 & \multicolumn{4}{c |}{E\&M}  & \multicolumn{4}{c |}{F\&R} \\
 Model & BLEU & Content & Fluency & Formality & BLEU & Content & Fluency & Formality\\ 
 \hline
 SimpleCopy & 50.28 & - & - & - & 51.66 & - & - & - \\
 Target & 99.99 & {5.54} & 4.79 & 2.31 & 100.00 & {5.54} & {4.79} & 2.30  \\
 \hline
 Rule-based & 60.37 & - & - & -  & 66.40 & - & - & - \\
 NMT \cite{shakespeare} & 68.41 & - & -  & - & 74.22 & - & - & - \\
 Transformer \cite{transformer} & 67.97 & - & - & - & 74.20 & - & - & - \\
 Hybrid Annotations ~\citep{formality}* & 69.63 & 5.22 & 4.62 & 1.97 & 74.43 & 5.29 & 4.53 & 2.04 \\ 
 NMT Multi-task \cite{nmtmultitask} & 72.13 & - & - & - & 75.37 & - & - & - \\
 Pretrained w/ Rules ~\citep{rules} & 72.70 & \textbf{5.38} & 4.51 & 1.67 & 76.87 & \textbf{5.64} & 4.63 & 1.78  \\
 \hline
 Dual Reinforcement** \citep{reinforcement} & - & - & - & - & 41.9 & - & - & - \\
 \hline
 Ours Base & 74.66 & 4.93 & 4.33 & 1.82 & 78.89 & 5.06 & 4.36 & 1.84 \\
 $\;$ w/ CNN discriminator* & 75.04 & - & - & - & 79.05 & - & - & - \\
 $\;$ w/ LM discriminator* & 75.65 & 5.33 & 4.69 & 2.30 & 79.50 & 5.35 & 4.66 & \textbf{2.31} \\
 $\;$ w/ LM* + MMI & 76.19 & - & - & -  & 79.92 & - & - & - \\
 Ours* & \textbf{76.52} &  5.35 & \textbf{4.81} & \textbf{2.38}  & \textbf{80.29} & 5.42 & \textbf{4.74} & \textbf{2.31} \\
 \hline
\end{tabular}
}
\caption{Results on GYAFC Dataset. An average of 3 runs was used for each model to calculate BLEU. Models with * leverage extra data via semi-supervised methods. ** represents unsupervised models. The description for the models is given in Section 4.3. The best scores  (besides the target) for each metric are in \textbf{bold}.}
\end{center}
\end{table*}

\subsection{Baselines and Model Variants}
We compared our approach with several baseline methods as follows:
\begin{itemize}
\item \textbf{SimpleCopy}: Simply copying the source sentence as the generated output.
\item \textbf{Target}: Human-generated outputs.
\item \textbf{Rule-based} ~\citep{grammarly}: Using hand-made rules.
\item \textbf{NMT} ~\citep{shakespeare}: A LSTM encoder-decoder model with attention.
\item \textbf{Transformer} ~\citep{transformer}: A Transformer architecture with the same configuration as our encoder and decoder.
\end{itemize}

We also compared our model with previous state-of-the-art works:
\begin{itemize}
\item \textbf{Hybrid Annotations} \citep{formality}: Uses CNN-based discriminator and cyclic reconstruction loss in a semi-supervised setting.
\item \textbf{NMT Multi-Task} \citep{nmtmultitask}: Solves two tasks: monolingual formality transfer and formality-sensitive machine translation jointly using multi-task learning.
\item \textbf{Pretrained w/ Rules} \citep{rules}: Uses a pre-trained OpenAI GPT-2 model and a combination of original and rule-based processed sentences to train the model.
\end{itemize}

The performances for these works were taken from the respective papers. We also introduced several variants of our model for comparison:
\begin{itemize}
\item \textbf{Ours Base}. Pretrained uni-directional auto-encoder architecture from BART ~\citep{bart} fine-tuned on our data.
\item \textbf{Ours w/ CNN Discriminator}: A CNN architecture with 3 layers used on the output of the decoder. The discriminant was trained with unsupervised class-labeled data.
\item \textbf{Ours w/ LM Discriminator}: Two transformer-based language models with 4 layers, used on the output of the decoder.
\item \textbf{Ours w/ LM + MMI}: Model trained with MMI objective and LM discriminator.
\item \textbf{Ours}: Ours Base model trained with LM discriminator, MMI objective,  and cyclic reconstruction loss.  
\end{itemize}

\begin{table*}
\begin{center}
\begin{tabular}{|c | l|}
\hline
Model & Sentence \\
\hline
Informal & fidy cent he is fine and musclar \\
Hybrid Annotations \citep{formality} & Fidy Cent is fine and Muslim. \\
Pretrained w/ rules \cite{rules} & Fidy Cent is a fine and musclar artist. \\
Ours & 50 Cent is fine and muscular. \\
Human-Annotation & 50 Cent is fine and muscular. \\
\hline
Informal &  Plus she is a cray ****. \\
Hybrid Annotations &  She is a clay. \\
Pretrained w/ rules & She is a cray ****. \\
Ours & She is not very nice. \\
Human-Annotation & Also, she is a mentally unstable woman. \\
\hline
Informal & So far i haven't heard that shes come back here (Arkansas)? \\
Hybrid Annotations & I have not heard that she is in Arkansas. \\
Pretrained w/ rules & So far, I have not heard that she is coming back here(Arkansas). \\
Ours & So far I have not heard that she has returned to Arkansas. \\
Human-Annotation & So far I have not heard that she returned to Arkansas. \\
\hline
\end{tabular}
\end{center}
\caption{Some sample outputs from various models.}
\end{table*}

\subsection{Results}
The results are summarized in Table 2. 
Compared to various baselines such as Pretrained w/ Rules \cite{rules}, our proposed models achieved significant improvements with 3.82 absolute increase of BLEU on E\&M and an increase of 3.42 on F\&R. 
By utilizing the language model discriminator and mutual information maximization, \texttt{Ours} achieved  state-of-the-art results on both subsets  of the GYAFC dataset in terms of BLEU, boosting the BLEU to 76.52 and 80.29  on E\&M and F\&R respectively. Our contributions increase the score by 2-3 points compared to the fine-tuned BART baseline as well. This validates the effectiveness of our semi-supervised formality style transfer models. Details on runtime and memory requirements can be found in Appendix. Our contributions increase the performance without increasing the test-time or memory requirements significantly.

Consistent with this quantitative result, 
human annotation results showed that \texttt{Ours} produced more fluent and more formal outputs compared to our selected baselines. \texttt{Pretrained w/ Rules} was rated to have better content preservation, but lower fluency and formality. This is possibly due to different approaches taken to deal with slang and idiomatic expressions in language, as described in Section 5.3 (Type 7). \citet{rules} tends to keep the content at the cost of formality of the output, while \citet{formality} and our model often ignore the content. For example, our model's output of ``\emph{the two boys rednecked as hell play guitar}'' is ``\emph{The two boys play guitar.}'', omitting details like ``red-neck'' which are rarely mentioned in formal language.

Moreover, we observed that there are comparable human annotation results between \texttt{Target} and \texttt{Ours}. Our model achieved  slightly higher scores on the formality of the sentences compared to  human-generated outputs. This may suggest that our model has a tendency to increase the formality of a sentence, even if it loses a bit of meaning preservation.
We also found that additional unsupervised data helps: compared to \texttt{Ours Base}, language model discriminator improves performance significantly (with BLEU scores from 74.66 to 75.65, and from 78.89 to 79.50). Note that our method is generic, and can be further combined with baseline methods, such as \citet{rules, nmtmultitask}.

We notice that BLEU does not necessarily correlate well with improved fluency, which is consistent with previous studies \citep{grammarly, bleuproblems}. Many fluent sentences did not capture the meaning of the sentence well, which reduces BLEU. Conversely, it is possible to have high intersection with the gold label sentence but still not be fluent.

Some qualitative results from our best-performing model (by BLEU score in Table 3), \citet{formality}, \citet{rules} and target sentences, are provided in Table 4. 
We observed that our model consistently generates better translations compared to the previous methods, especially in terms of dealing with proper nouns, informal phrases and grammatical mistakes.

\subsection{Testing on Unsupervised data}
\begin{table*}
\begin{center}
{\begin{tabular}{|l|c c c| c c c |}
\hline
& \multicolumn{3}{c |}{Yelp} & \multicolumn{3}{c |}{Amazon} \\
Model & ACC & BLEU & G-Score & ACC & BLEU & G-Score\\ 
\hline
SimpleCopy & 2.4 & 18.0 & 6.57 & 18.9 & 39.2 & 27.2 \\
Target & 69.6 & 100.0 & 83.4 & 41.3 & 99.9 & 64.2  \\
\hline
Cross Aligned \citep{disc1} & 73.7 & 3.1 & 15.1 & 74.1 & 0.4 & 5.4  \\
Style Embedding \citep{unsupbase} & 8.7 & 11.8 & 10.1  & 43.3 & 10.0 & 20.8 \\
Multi Decoding \citep{unsupbase} & 47.6 & 7.1 & 18.4  & 68.3 & 5.0 & 18.5 \\
Template Based \citep{delete} &  81.7 & 11.8 & 31.0 & 68.7 & 27.1 & 43.1 \\
Retrieve Only ~\citep{delete} & \textbf{95.4} & 0.4 & 6.2 & 70.3 & 0.9 & 8.0 \\
Delete Only ~\citep{delete} & 85.7 & 7.5 & 25.4 & 45.6 & 24.6 & 33.5 \\
Delete \& Retrieve ~\citep{delete} & 88.7 & 8.4 & 27.3 & 48.0 & 22.8 & 33.1 \\
Dual Reinforcement ~\citep{reinforcement} & 85.6 & 13.9 & 34.5 & - & - & - \\
Word-level Conditional GAN ~\citep{conditionslgan} & 87.8 & 9.6 & 29.1 & \textbf{77.4} & 6.7 & 22.8 \\
Iterative Matching ~\citep{iterative} & 87.9 & 4.3 & 19.4 & - & - & - \\
Ours & 86.2 & \textbf{14.1} & \textbf{34.9} & 68.9 & \textbf{28.6} & \textbf{44.4} \\
\hline
\end{tabular}
}
\caption{Results on Sentiment Transfer datasets. Results were averaged across two directions: negative-to-positive and positive-to-negative sentiment transfer. An average of three runs was used for each directions. Here ACC and GM mean Accuracy and G-Score respectively. The best scores (besides the target) for each metric are in \textbf{bold}.}
\end{center}
\end{table*}
We further extended our method to unsupervised tasks, using only cyclic reconstruction and Language Discriminator losses as our training objective. Sentiment Transfer corpus \citep{delete} from Yelp and Amazon was used for evaluation. The statistics are given in Table 2. The corpora include separate negative and positive sentiment data without parallel data. 
We followed the evaluation protocol and baselines from ~\citet{delete}.
In addition to BLEU, we used two additional metrics for evaluation: 
(1) \textbf{Accuracy}: The percentage of sentences successfully translated into positive, as measured by a separate pre-trained classifier. 
(2) \textbf{G-Score}: The geometric Mean of accuracy and BLEU scores. We rank our models by G-Score, following ~\citet{gscore}, since there is a trade-off between accuracy and BLEU, as changing more words can get better accuracy but lower content preservation.

We used the script and sentiment classifier from ~\citet{delete} to evaluate our outputs. Results were averaged for the two directions: positive-to-negative sentiment transfer and negative-to-positive sentiment transfer, with 500 sentences in the test set for each direction.

We compared our results with previous state-of-the-art approaches. Style Embedding and Multi Decoding \citep{unsupbase} learn an embedding of the source sentence such that a decoder can use it to reconstruct the sentence, but a discriminator, which tries to identify the source attribute using this encoding, fails. Cross-Aligned \citep{disc1} also encodes the source sentence into a vector, but the discriminator looks at the hidden states of the RNN decoder.

\citet{delete} extract content words by deleting style-related phrases, retrieves relevant target-related phrases and combines them using a neural model. They provide three variants of their model. Word-level Conditional GAN ~\citep{conditionslgan} also tries to separate content and style with a word-level conditional architecture. Dual Reinforcement \citep{reinforcement} uses reinforcement learning for bidirectional translation without separating style and content. Iterative Matching \citep{iterative} iteratively refines imperfections in the alignment of semantically similar sentences from the source and target dataset. We used the performance numbers for these approaches from either the original papers when the evaluation protocol is similar to ours or by evaluating publicly released outputs of the models.

We achieved state-of-the-art results on both Yelp and Amazon Sentiment Transfer corpus, as shown in Table 5. Our model attains slightly lower accuracy on sentiment classification of output sentences, but preserves more content compared to previous models, resulting in the highest G-Score on both datasets. This suggests that our approach can generalize well to unsupervised style transfer tasks.

\section{Model Analysis and Discussion}
Although our model performed well on formality style transfer, there is still a gap compared to human performance.  To understand why the task is challenging and how future research could advance this direction, we take a closer look at formality dataset, model generation errors, and certain challenges that existing approaches struggle with.
\begin{figure}[t]
\includegraphics[width=1\linewidth]{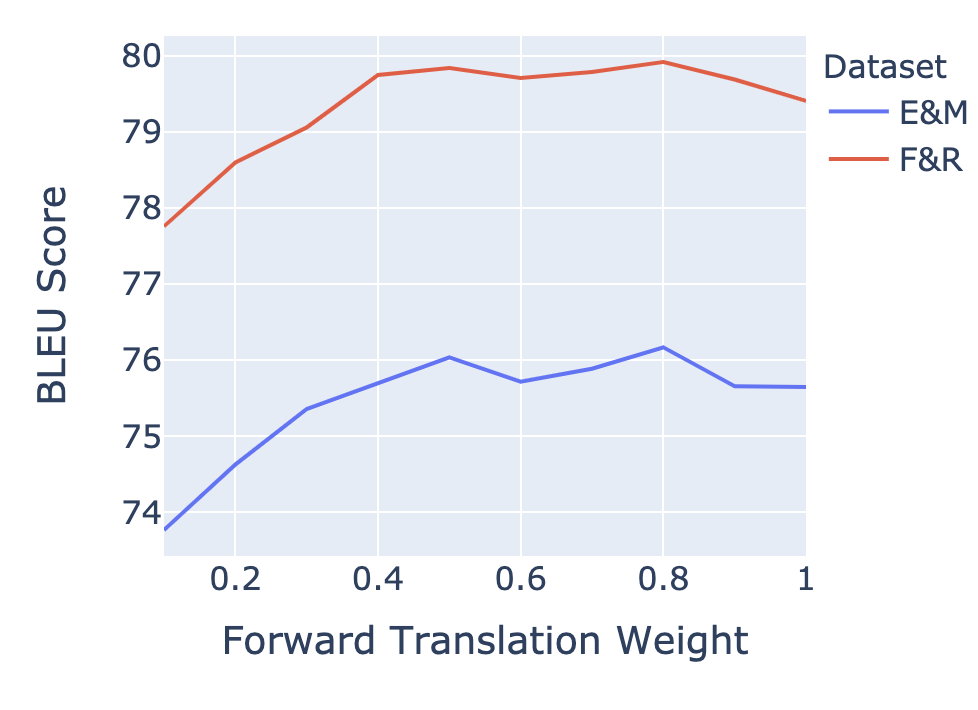}
\caption{Performance with forward translation weight}
\end{figure}
\subsection{Effect of Forward Translation Weight}
As mentioned in Section 3.3, MMI objective is equivalent to a weighted sum of source-to-target and target-to-source translation. We show the effect of forward translation weight, $\lambda$ in Figure 2, and find that using MMI objetive helps performance as compared to baseline translation loss (which corresponds to $\lambda = 1.0$). 
However, equivalent weighing of the two directions (corresponding to $\lambda = 0.5$) does not result in the best performance: a bias towards the informal to formal direction($\lambda = 0.8$) gives better BLEU scores. We posit that this could be because unlike formal sentences, informal sentences do not follow a particular style: they vary from structurally correct with some mistakes to just a collection of telegram-style keywords, and hence the objective of generating this should be assigned less importance than the forward task.
\subsection{Cyclic and Discriminator Loss}
In our model, we used unsupervised class labeled data to train our model using cyclic and discriminator loss. We also conducted experiments to use these losses for parallel data as well. However, training on parallel data using these objectives in addition to MMI objective did not result in additional improvements, while increasing the training time and memory requirements. Partially, this could be because maximizing target sentence probability already captures the target style, hence discriminator loss does not help. Similarly, maximizing Mutual Information ensures that target-to-source translation is also a maximisation objective during training, hence reducing the effectiveness of cyclic reconstruction loss. Therefore, we concluded that maximizing mutual information during training is sufficient for parallel data.

\subsection{Challenges in Formality Text Transfer}
We conduct a thorough examination of the GYAFC dataset and categorize the challenges into the following categories:
\begin{enumerate}
\item \textbf{Informal Phrases and Abbreviations}: Presence of ``informal'' phrases (\emph{what the hell}), emojis (\_:)) and abbreviations (\emph{omg, brb}).

\item \textbf{Missing Context}: A lack of context of the conversation (for example, \emph{``It had to be the chickin''}) or a lack of punctuation or proper capitalization cues (\emph{``can play truth or dare or snake and ladders''}).

\item \textbf{Named Entities}: Proper nouns and popular references like ``Fifty Cent'' or ``eBay'' should not be changed despite the wrong pluralization and capitalization, respectively. This is worsened by the lack of any capitalization or punctuation cues to find named entities.

\item \textbf{Sarcasm and Rhetorical Questions}: Rhetorical questions, sarcastic language and negations have been long-standing problems in NLP \citep{negation}. For example, ``\emph{sure, because this is so easy}'' is sarcastic and should not be translated literally. 

\item \textbf{Repetition}: Informal text often has a lot of redundant information. For example, ``\emph{I used to work at the store and met him while i was working there.}'' can be formally structured as ``\emph{I met him while i was working at the store.}''.

\item \textbf{Spellings and Grammar Errors}: This is prevalent in most ($>$60\%) informal sentences.

\item \textbf{Slang and Idiomatic Expressions}: Some sentences have words especially nouns, adjectives and adverbs that can be considered as slang, idiom, and even discriminatory.
\end{enumerate}

\begin{table}
\begin{center}
\resizebox{!}{5em} {\begin{tabular}{|c|c| c| c|}
\hline
Type & Input (\%) & Output (\%) & Resolved (\%)\\ 
\hline
1 & 16 & 7 & 56 \\
2 & 5 & 4 & 20  \\
3 & 12 & 3 & 75 \\
4 & 2 & 2 & 0\\
5 & 3 & 0 & 100 \\
6 & 61 & 7 & 89 \\
7 & 5 & 0 & 100 \\
\hline
\end{tabular}
}
\caption{The breakdown of challenge types for formality style transfer, and their percentages in the source \emph{Input},  generated \emph{Output}, and the percentage of challenges successfully resolved by our model.}
\end{center}
\end{table}

We randomly sampled 100 sentences from the dataset to estimate the prevalence of such challenges. We also examined the output from our model to analyze if a challenge has been solved or still presents an issue to the model. The result is summarized in Table 6.
We found that our models resolved most spelling and grammatical mistakes (Type 6), and performs well with avoiding repetition (Type 5). However, missing context, informal expressions and named entities continue to be challenging. One major challenge is the inability to correct sarcastic/rhetorical sentences (Type 4).

\section{Conclusion}
This work introduces a semi-supervised formality style transfer model that utilizes both a language model based discriminator to maximize the likelihood of the output sentences being formal, and a mutual information maximization loss during training. 
Experiments conducted on a large-scale formality corpus showed that our simple method significantly outperformed previous approaches in terms of both automatic metrics and human judgement. 
We also demonstrated that our model can be generalized well to unsupervised style transfer tasks.  We also discussed specific challenges that current approaches faced with this task.

\section*{Acknowledgements}
We thank the reviewers and members of Georgia Tech SALT group for their feedback on this work. We acknowledge the support of NVIDIA Corporation with the donation of GPU used for this work. 

\bibliography{anthology,acl2020}
\bibliographystyle{acl_natbib}

\end{document}


\maketitle
\appendix

\begin{table*}
\begin{center}
\resizebox{1 \textwidth}{!}
{\begin{tabular}{|l|c c c c c|}
\hline
Model & \#param(M) & Training time(min)& Testing time(ms) & Memory(MB) & Valid ppl\\ 
\hline
Base & 406 & 95 & 14 & 8203 & 3.54 \\
CNN Discriminator & 406 & 112 & 14 & 8542 & 3.53 \\
LM Discriminator & 495 & 129 & 14 & 9676 & 3.44 \\
LM + MMI & 495 & 133 & 14 & 9833 & 3.36 \\
LM + MMI + Cyclic Loss & 495 & 147 & 14 & 10296 & 3.32 \\
\hline
\end{tabular}
}
\caption{Computational requirements of our model. \emph{\#param} is the number of trainable parameters, \emph{Training time} is in minutes/epoch, \emph{testing time} is in milliseconds/sentence, and \emph{Valid ppl} is validation perplexity on GYAFC E\&M.}
\end{center}
\end{table*}

\section{Dataset Collection}
\subsection{BookCorpus formal data}
To collect formal data, we used the code from \citet{bookcorpuscode}, which collects data from smashwords \citep{smashwords}, which is the original source of BookCorpus. \citep{bookcorpus}. The post-processing was done similarly to \citet{bookcorpuscode}. We split each paragraph of the books into its constitutent sentences, and collected more than 1 million sentences from over 100 books, and randomly chose 20\% of the sentences for our training. We further removed all sentences with less than 5 or more than 20 tokens; and also removed all sentences with quotes, so as to remove dialogues, which could be of informal style.

\subsection{Twitter Informal data}
For informal data, we used Twitter data from \cite{twitterdataset}. The dataset contains more than 2 million tweets posted during October 2010. To pre-process the tweets, we collected the sentences from each tweet and removed URLs, hashtags and Twitter username references. Similar to BookCorpus, we discarded all sentences with fewer than 5 or greater than 20 tokens.

\section{Hyperparameter Details}
For most of our parameters, we used the same values as \citet{fairseqbart}. Total number of epochs was set to 30, including 10 for pre-training the discriminator. The number of max tokens per batch was set to 64, and update frequency was 4. The following hyperparameters were tuned based on perplexity of translation of validation set:
\begin{enumerate}
\setlength \itemsep{-0.25em}
    \item \textbf{Forward-translation weight, $\lambda$}: This decides the relative weight of forward translation to backward-translation, as described in Section 3.3. It was tuned in the range of [0.1,1.0] with 0.1 intervals.
    \item \textbf{Cyclic Reconstruction loss weight, $w_{cycle}$}: This decides weight of cyclic loss for the training, as described in Section 3.4. It was tuned in the range of [0.2, 1.4] with 0.2 intervals. Best performance was achieved for $w_{cycle} = 0.6$ in semi-supervised setting (for GYAFC) and $w_{cycle}=1.0$ in unsupervised setting (for Sentiment Transfer).
    \item \textbf{Discriminator loss weight, $w_{disc}$}: This decides weight of discriminator loss for the training, as described in Section 3.4. It was also tuned in the range of [0.2, 1.4] with 0.2 intervals. The final values selected was $w_{disc} = 1.0$ for both semi-supervised and supervised settings.
    \item \textbf{Learning rate}: Learning rate was tuned in the range $[e^{-8},e^{-2}]$, and experiments were run for each order of magnitude. The learning rate selected differed depending on datasets, settings, and values of other hyperparameters.
\end{enumerate}

\section{Computational Requirements}
The runtime and memory requirements of various variations of our model are given in Table 1. The models are as described in Section 4.3. Validation perplexity is calculated on Entertainment and Music (E\&M) domain of Grammarly Yahoo Corpus (GYAFC) \cite{grammarly}  Dataset. We notice that our contributions increase the performance without increasing the memory, number of parameters or testing time significantly.

\section{Human Evaluation Details}
Workers were hired from Amazon Mechanical Turk for human evaluation of the model outputs. Three workers per sentence were asked to rate each style transfer on meaning preservation, fluency and formality, as described in Section 4.2. The interface shown to workers is given in Figure 1.

\begin{figure*}
\includegraphics[width=1\linewidth]{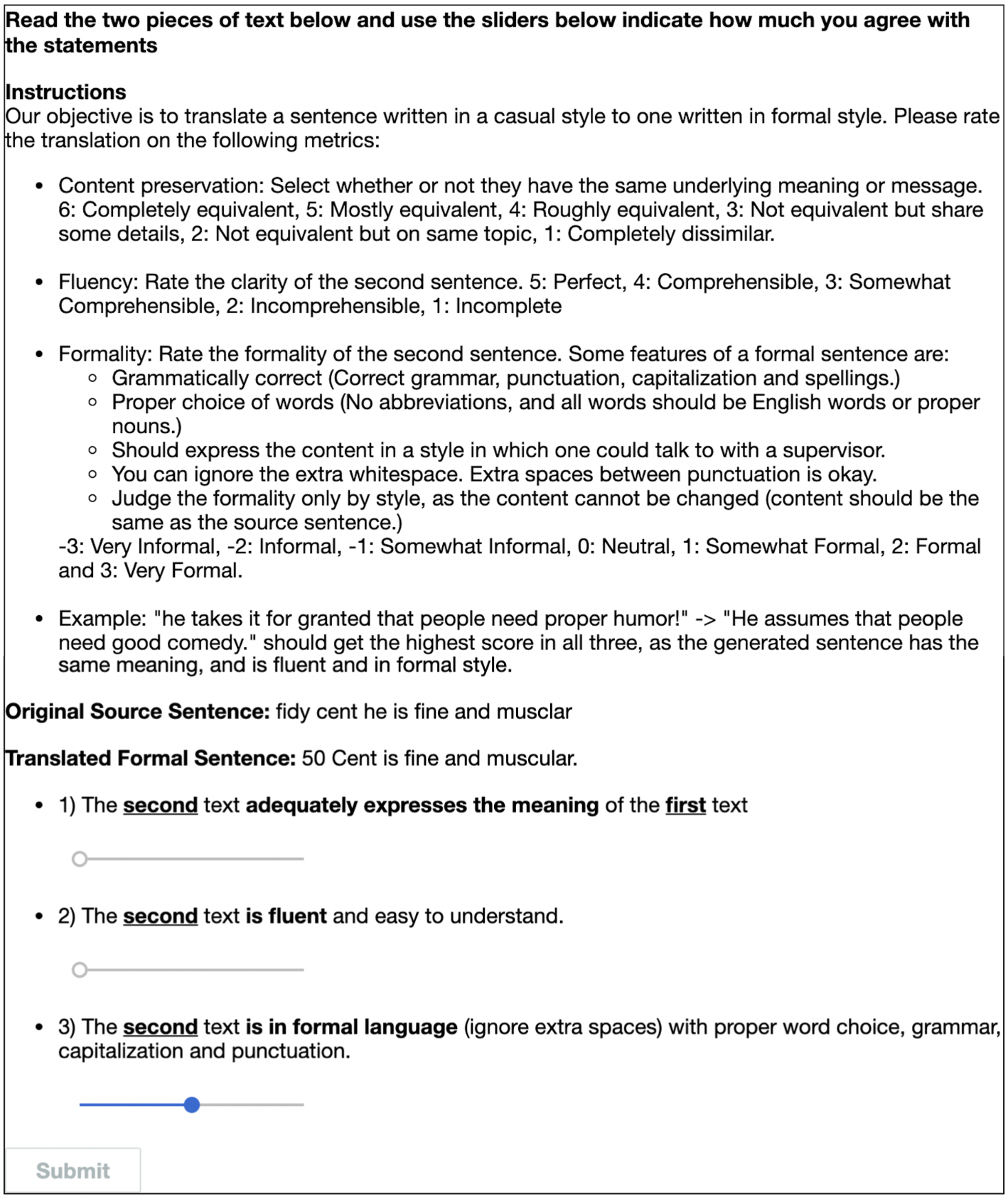}
\caption{User interface for human evaluation, as seen by an Amazon Mechanical Turk worker.}
\end{figure*}

\bibliography{appendix}
\bibliographystyle{acl_natbib}